\documentclass[twoside]{article}
\usepackage{svg-extract}

\usepackage{aistats2024}
\usepackage{amsmath, amsthm, amssymb, bm, xcolor, verbatim,svg, subcaption}
\usepackage{tabularx, array, booktabs}
\usepackage{natbib}
\usepackage{enumitem}
\setlist{topsep=0pt, leftmargin=*}
\usepackage{titlesec}
\titlespacing\subsection{0pt}{0pt plus 4pt minus 2pt}{0pt plus 2pt minus 2pt}
\usepackage{algorithm}
\usepackage{algpseudocode}

\DeclareMathOperator*{\argmax}{arg\,max}

\newcommand{\rudycomment}[1]{}
\newcommand{\br}[1]{\left({#1}\right)}
\newcommand{\norm}[1]{\left\|{#1}\right\|}
\newcommand{\x}{{\bf x}}
\newcommand{\y}{{\bf y}}
\newcommand{\z}{{\bf z}}

\newcommand{\xb}{\mathbf{x}}
\newcommand{\alphab}{\bm{\alpha}}
\newcommand{\cb}{\mathbf{c}}
\newcommand{\thetab}{\bm{\theta}}
\newcommand{\lzerodomain}{\mathbb{X}^l_{\epsilon, 1}}

\DeclareMathOperator{\sign}{sign}

\begin{document}

%

%

\twocolumn[
\aistatstitle{Verified Neural Compressed Sensing}
\aistatsauthor{Rudy Bunel$^*$ \\\
Google DeepMind \\
\texttt{rbunel@google.com} \\
\And
Krishnamurthy (Dj) Dvijotham$^*$ \\
Google DeepMind \\
\texttt{dvij@google.com}\\
\And
M. Pawan Kumar $^*$ \\
Google Deepmind \\
\texttt{mpawan@google.com}\\
\AND
Alessandro De Palma\\
Inria, École Normale Supérieure, PSL University, CNRS\\
\texttt{alessandro.de-palma@inria.fr}\\
\And
Robert Stanforth\\
Google Deepmind\\
\texttt{stanforth@google.com}\\
}
]

\begin{abstract}
We develop the first (to the best of our knowledge) provably correct neural networks for a precise computational task, with the proof of correctness generated by an automated verification algorithm without any human input. Prior work on neural network verification has focused on partial specifications that, even when satisfied, are not sufficient to ensure that a neural network never makes errors. We focus on applying neural network verification to computational tasks with a precise notion of correctness, where a verifiably correct neural network provably solves the task at hand with no caveats. In particular, we develop an approach to train and verify the first provably correct neural networks for compressed sensing, i.e., recovering sparse vectors from a number of measurements smaller than the dimension of the vector. We show that for modest problem dimensions (up to $50$), we can train neural networks that provably recover a sparse vector from linear and binarized linear measurements. Furthermore, we show that the complexity of the network (number of neurons/layers) can be adapted to the problem difficulty and solve problems where traditional compressed sensing methods are not known to provably work. 
\end{abstract}

\section{Introduction}
Neural network verification techniques developed over the years have shown great promise in the ability to certify that neural networks satisfy desirable properties like robustness to adversarial examples \citep{cohen2019certified, wong2018provable}, fairness \citep{ruoss2020learning}, Lipschitzness \citep{wang2023direct}. However, these examples are all of the form that only establish a partial specification. Just because a neural network satisfies this specification, it does not mean that it will output the correct response for all inputs. In fact, for most prediction tasks, even defining the correct response for every input is challenging, and in practice, there are natural trade-offs between the ability of a network to satisfy various specifications, say accuracy and fairness, or accuracy and robustness. \looseness=-1

Motivated by this, we consider use-cases where neural networks are applied to a mathematical or scientific problem with a precise notion of correctness. In such cases, if we were able to prove that the network we learn satisfies the precise notion of correctness, we clearly have solved the problem at hand. In recent years, there have been several applications of neural networks towards solving mathematical or computational problems like solving PDEs \citep{karniadakis2021physics},  sparse signal recovery \citep{bora2017compressed} and even sorting \citep{mankowitz2023faster}. Some prior works achieve correctness by construction since they only use neural networks as components of a provably correct solver (like a SAT or mixed integer program sovler) \citep{selsam2018learning}, and others provide human generated custom correctness proofs for specific neural network based algorithms \citep{bora2017compressed}. However, a neural network trained directly to solve a mathematical or scientific problem without any specific architectural choices to ensure correctness by construction, has not been proven to be correct thus far using an automated verification algorithm.

In this work, we take initial steps towards the objective of training and verifying neural networks that provably solves a concrete mathematical task, that of compressive sensing or recovering a sparse vector from a small number of measurements. Neural networks have previously been applied to this problem (for example, in \citep{shi2019scalable}) where it was shown that neural networks can achieve a stronger performance than classical non-neural compressed sensing algorithms. However, no prior work has shown that the neural networks are provably correct, in the sense that they correctly recover every sparse vector. We list our specific contributions in the following section and provide a thorough development in subsequent sections.

\subsection{Contributions}
We study the recovery of a vector $\x$ from measurements $\y$ where $\y$ is a function of $\x$ (often linear) with $\y$ being lower dimensional than $\x$. Informally (Precise mathematical definitions are deferred to Section \ref{sec:prelim}), we train a neural network $\text{NN}$ that satisfies
\begin{align}
    \text{NN}\br{\y}=\x \quad \forall \text{ sparse } \x  \label{eq:NNverify} 
\end{align}
 Relative to this setting, our contributions can be summarized as follows:

\begin{itemize}
    \item We formalize the problem of proving that \eqref{eq:NNverify} is true as a neural network verification problem \citep{katz2017reluplex} and adapt verification algorithms to solve this.
    \item We develop a variant of adversarial training \citep{madry2017towards} that trains neural networks towards satisfying \eqref{eq:NNverify} and adapt techniques that make the resulting neural networks easier to verify \citep{depalma2023expressive} to this setting.
    \item We show that our provably correct neural networks have several advantages over traditional compressed sensing algorithms with provable guarantees \citep{donoho2006compressed}: 1. They can adapt their complexity (number of neurons/number of layers) to the complexity of the compressed sensing problem being solved, by training a neural network specific to a particular dimension of $\x, \y$. 2. They can extend (while remaining provably correct) to situations where traditional algorithms lack provable guarantees, for example, when the measurements are a mixture of binarized and linear. \looseness=-1
    \item We conduct exhaustive experiments to study the impact of various factors like the dimensions of $\x, \y$ and the minimum non-zero entry in $\x$ on the complexity of training and verification. 
\end{itemize}

\section{Related Work}

{\bf Provable learning based compressed sensing:} \citet{bora2017compressed} introduce a method for compressed sensing based on a generative model, where the vector being recovered is assumed to be of the form $G\br{\z}$ where $\z$ is a latent vector of lower dimension and $G$ is a generative model that maps lower dimensional latents to a higher dimensional space. In \citep{rout2023solving}, the authors tackle linear inverse problems similar to the ones from this paper using diffusion models. Our work is distinct from these works in two ways: 1. We do not develop an custom architecture for the task, we simply train a neural network (MLP with ReLU activations) directly to predict the support of the sparse vector being recovered. 2. We use automated neural network verification techniques to prove correctness, so our techniques automatically generalize to harder tasks like mixed binary and linear measurements, or other neural network architectures. 3. The provable guarantees in \citep{bora2017compressed} depend on being able to solve certain optimization problems exactly (in particular the ability to compute $\min_{\z} \norm{G\br{\z}-\x}$), which are themselves NP-hard and hard to solve in a provably optimal manner in general.

{\bf Prior work on neural network verification:} We build on both state of the art neural network verification techniques \citep{wang2021beta} and techniques for training neural networks that are easier to verify \citep{depalma2023expressive}, but adapt it to the setting of verifying that \eqref{eq:NNverify} is true.

{\bf Prior work on neural solvers:} Neural approaches have been developed to discover novel approaches to mathematical or computational problems including matrix multiplication \citep{fawzi2022discovering} and novel instruction-level sort algorithms \citep{mankowitz2023faster}. However, in all these cases, the provably correct algorithms are a result of specific modeling choices and are baked in to the learning process. In contrast, our work allows a search over neural networks that include correct and incorrect decoders for compressed sensing, and our training and verification algorithms target learning a network that can be verified to be correct. In this sense, our framework is more flexible as it can extend to general tasks where it may be difficult to provide an efficient correct by construction solution, like compressed sensing.

\section{Preliminaries}\label{sec:prelim}
\paragraph{Compressed Sensing of Sparse Signals.}
Compressed sensing refers to the problem of recovering a sparse signal from its linear measurements. Formally, let $A \in \mathbb{R}^{m \times n}$ be a sensing or measurement matrix, where $m < n$. Our goal is to recover ${\bf x}$ from the knowledge of ${\bf y} = A{\bf x}$. As $m < n$, at first glance the problem may appear to be ill-posed. Nonetheless, it can be shown that if ${\bf x}$ is assumed to be sparse, then there exists a unique solution. Note, however, that the existence of a unique solution does not imply the existence of an efficient algorithm to recover this solution.

Formally, compressed sensing can be formulated as the following constrained optimization problem:
\begin{equation}
\min_{\bf x} ||{\bf x}||_0 \mbox{ subject to } {\bf y} = A{\bf x}.
\label{eq:compressedsensing}
\end{equation}
Here, $||\cdot||_0$ refers to the $\ell_0$ pseudo-norm, that is, the number of non-zero elements of the vector. The above problem is known to be NP-hard, with the source of its hardness lying in the non-convexity of the $\ell_0$ pseudo-norm. In this work, we additionally assume that all the elements of {\bf x} are less than or equal to $1$. This can be easily incorporated into the above formulation via the convex constraints $x_i \leq 1$ for all elements $x_i$ of {\bf x}. Furthermore, we also assume that if an element of {\bf x} is not zero, then its value is at least $\epsilon$. While this assumption would add further non-convexity to the above formulation (for example, by introducing constraints $x_i(x_i - \epsilon) \geq 0$), as will be seen later, it can be easily incorporated within our novel approach.

\paragraph{Compressed Sensing with Binary Measurements.}
We also consider a modification of compressed sensing where some of the measurements are binary. In other words, while the exact values of the measurements is unknown, we know whether they are greater than a pre-specified threshold or not. Formally, we are provided with two types of measurements, ${\bf y}_1 = A_1{\bf x}$ and ${\bf y}_2 = \sign(A_2{\bf x} - \tau)$ where $A_1 \in \mathbb{R}^{m_1 \times n}$, $A_2 \in \mathbb{R}^{m_2 \times n}$ and the operator `$\sign$' returns the sign of its arguments (that is, +1 if greater than or equal to 0 and -1 otherwise). The problem of compressed sensing with binary measurements can be formulated as the following optimization problem:
\begin{equation}
\min_{\bf x} ||{\bf x}||_0 \mbox{ s.t. } \ {\bf y}_1 = A_1{\bf x}, \mbox{ and } {\bf y}_2 \odot (A_2{\bf x} - \tau) \geq 0,
\label{eq:compressedsensingbinary}
\end{equation}
where $\odot$ refers to elementwise multiplication. Unlike in the previous setting, the use of binary measurement implies that the solution may not be unique. Instead, the goal in this setting is to identify the correct support for the underlying sparse vector {\bf x}.

\paragraph{Traditional Solvers for Compressed Sensing.}
Several algorithms have been designed for the compressed sensing problems discussed above. Among the most popular of these belong to the family of convex relaxations, where the original non-convex problem is replaced by a convex approximation. As a concrete example, consider the following convex relaxation of problem~(\ref{eq:compressedsensing}):
\begin{equation}
\min_{\bf x} ||{\bf x}||_1 \quad \mbox{ s.t. } {\bf y} = A{\bf x}.
\label{eq:basispursuit}
\end{equation}
The above problem can be formulated as a linear program, thereby lending itself to polynomial time solvers.

The resulting approach, termed basis pursuit, has been widely analysed, leading to a deep understanding of conditions under which the relaxation is tight~\citep{candes2006near, donoho2003optimally}. However, such an approach is onerous. Specifically, for each variation of compressed sensing (for example, those with binary measurements) we would be required to design its own bespoke relaxation, together with a non-trivial analysis to establish its correctness.

\section{Our Approach}
Unlike the traditional approach to algorithm design, we wish to synthesize a deep neural network that can be trained to adapt itself to a given setting of compressed sensing. In order to be more conducive to a neural solver, we address an equivalent problem to~(\ref{eq:compressedsensing}) and~(\ref{eq:compressedsensingbinary}), where we are given the desired $\ell_0$ norm $l$ of the underlying signal and wish to recover its support.
In more detail, for compressed sensing of sparse signals, the input to the neural network is the linear projection {\bf y} of the signal {\bf x} (see equation~(\ref{eq:compressedsensing})). The output of the neural network is of the same size as ${\bf x}$ where each element is a logit indicating whether the corresponding element of ${\bf x}$ is non-zero. By choosing the elements whose logits are positive, we recover the support of the signal, which in turn provides sufficient information to recover the signal itself by solving a linear system of equations. For compressed sensing with binary measurements, the input to the neural network are both ${\bf y}_1$ and ${\bf y}_2$ (see equation~(\ref{eq:compressedsensingbinary})). As mentioned earlier, when some of the measurements are binary, the solution for the inverse problem is not unique. Instead, the neural network output provides the linear subspace that is consistent with the measurements provided.

Recall that the neural network that we wish to synthesise should not only be empirically accurate, but provably so. We achieve this goal via two steps. First, we estimate the parameters of the neural network via adversarial training. Second, we establish its correctness via neural network verification. In what follows, we first describe the neural network used in our work, followed by a description of each of the two steps.

\subsection{Model}
While (adversarial) training of neural networks has been successfully scaled to very large models with billions of parameters, its verification remains computationally challenging. In order to ensure that we can establish its correctness, we would like to choose a simple model that is still powerful enough to solve the compressed sensing task.

The neural network that we found to be suitable for our purpose starts by computing several scaled versions of the measurements and concatenating them together to obtain a high dimensional vector. In other words, we apply a fixed linear transformation to each element of the measurement individually to obtain an intermediate representation. This is then followed by a set of $h$ fully connected hidden layers with ReLU non-linearities, each of which have a skip connection with the first layer. Empirically, we found the skip connections to be critical to the correctness of the neural network in all but the easiest settings reported in our experiments. The choice of the non-linearity is driven by the fact that verifying networks with ReLUs can be formulated as a mixed integer linear program, and can therefore be addressed via a branch-and-bound approach~\citep{bunel2018unified}. Finally, the last layer provides an $n$-dimensional vector, whose $i$-th element corresponds to the logit for the $i$-th element of the underlying sparse signal ${\bf x}$. We denote the parameters of the network by $\bm{\theta}$ and its output by ${\bf z}({\bf x}; \bm{\theta})$.

\subsection{Adversarial Training}
\label{sub:at_for_pareps}
The main advantage of a neural solver is that it can be easily adapted to any setting of compressed sensing during training. In particular, recall that in our case we know that each signal of interest ${\bf x}$ satisfies $||{\bf x}||_0 = l$ and $x_i \in \{0\} \cup [\epsilon, 1]$. We denote the set of all such signals by $\lzerodomain$. These restrictions add to the complexity of traditional approach by introducing non-convex constraints to the mathematical optimization problem. However, in our case, it is simply a matter of generating training data that complies with these restrictions---a significantly easier task.

In more detail, given an $n$-dimensional signal {\bf x}, we denote its support as an $n$-dimensional vector $s({\bf x})$. In other words, the $i$-th element of $s({\bf x})$, denoted by $s_i({\bf x})$, is $0$ if $x_i$ is $0$ and is $1$ otherwise. Ideally, the support of the signal {\bf x} must match the value of the logit outputted by the neural network. In other words, $z_i({\bf x}; \bm{\theta})$ should be positive if $s_i({\bf x}) = 1$ and negative otherwise, thereby providing us with the correct support for {\bf x}.
For a given signal ${\bf x} \in \mathbb{X}^l_{\epsilon, 1}$ we define the mean binary cross-entropy loss with respect to the network parameters as \looseness=-1
\begin{equation}
l({\bf x}; \mathbf{\theta}) = \frac{1}{n} \sum_{i=1}^n -s_i({\bf x})z_i({\bf x}; \bm{\theta}) \log(1 + e^{z_i({\bf x}; \bm{\theta})}).
\end{equation}
Ideally, we would like to synthesize the neural network by solving the following problem:
\begin{equation}
    \min_{\bm{\theta}} \max_{{\bf x} \in \mathbb{X}^l_{\epsilon, 1}}  l({\bf x}; \mathbf{\theta}) + R(\bm{\theta}),
\end{equation}
where 
$R(\cdot)$ is a regularization term for the neural network parameters. Since our goal is to synthesize a model that is amenable to verification, we use IBP-R regularization~\citep{depalma2023expressive}. The loss term in the above objective is measured with respect to the worst-case signal.

While the above problem intuitively captures our desired network, there are two main sources of difficulty in solving it. First, the outer minimization with respect to $\bm{\theta}$ is highly non-convex. As is common in deep learning, we iteratively update the network parameters via Adam~\citep{KingBa15}.
Second, the inner maximization with respect to ${\bf x}$ is highly non-convex. We address this issue using two heuristics. First, we notice that empirically the hardest samples tend to be corners of the set $\lzerodomain$. In other words, they are vectors with $\ell_0$ norm $l$ whose non-zero elements are either $\epsilon$ or $1$. We exploit this observation by randomly sampling corners during each iteration of training. Second, as is common in the adversarial training literature, we approximately maximize the objective over $\lzerodomain$ via AutoPGD with multiple random initializations~\citep{croce2020}. It is worth noting that the projection to $\mathbb{X}^l_{\epsilon, 1}$, which is required to operationalise AutoPGD is computationally efficient. Specifically, we first clip all the values of a given signal ${\bf x}'$ to lie between $\epsilon$ and $1$, and then set all but the largest $l$ elements to $0$. The overall training approach is described in Algorithm~\ref{alg:train}.
\begin{algorithm}
\caption{Training the neural solver for compressed sensing.}
\label{alg:train}
\begin{algorithmic}[1]

\Require Initial network parameters $\bm{\theta}^0$.
\Ensure Final network parameters $\bm{\theta}^T$.
\State Set iteration counter $t = 0$.
\While{$t < T$}
\State $t = t+1$
\State Compute ${\bf g}^t = \nabla R(\bm{\theta}^t)$.
\State Sample a random corner ${\bf x}_c^t \in \lzerodomain$.
\State Obtain a sample ${\bf x}_a^t \in \lzerodomain$ using AutoPGD~\citep{croce2020}.
\State Choose ${\bf x}^t = \argmax_{{\bf x} \in \{{\bf x}_c^t, {\bf x}_a^t\}} l({\bf x}; \bm{\theta}^t)$.
\State Compute the gradient ${\bf g}^t = \nabla l({\bf x}^t; \bm{\theta}^t) + R(\bm{\theta}^t)$.
\State Update $\bm{\theta}^{t+1}$ using the current parameters $\bm{\theta}^t$ and all the gradients computed thus far via Adam~\citep{KingBa15}.
\EndWhile
\end{algorithmic}
\end{algorithm}

In addition to the network parameters $\bm{\theta}$, we may wish to learn the sensing matrix itself. This can be operationalised easily within our framework by computing the gradient of the objective function with respect to the measurement matrix and updating it similar to the network parameters. We hypothesize that doing so might result in a compressed sensing setting that can be more efficiently verified via its neural network based solver compared to the setting with a given fixed sensing matrix. In our experiments, we show that the above hypothesis indeed holds true by allowing for greater regularization when jointly estimating the sensing matrix and the neural network parameters.

\subsection{Formulation of the Proof of Correctness}
\label{sub:verif_formulation}
Once the network has been trained, we ensure that the support will be properly identified.
For every valid signal, we prove that for each coordinate, the corresponding logit will have the proper sign. 
\begin{subequations}
    \begin{align}
        \forall i \in [1, n], \ & \{\xb \in \lzerodomain, x_i \geq \epsilon\} \implies z_i(\xb, \thetab) > 0 \label{eq:verif_form_pos}\\
        & \{\xb \in \lzerodomain, x_i = 0\}  \implies z_i(\xb, \thetab) < 0\label{eq:verif_form_neg}
    \end{align}\label{eq:verif_props}
\end{subequations}
Methods from the neural network verification literature~\citep{auto_lirpa} allow to compute bounds on the output of the network, given a bounded input domain. 
Using those, if we can prove that a lower bound on the output of the network $z_i(\xb, \theta_b)$ is greater than 0, it is sufficient to show that the implication \ref{eq:verif_form_pos} holds.
We will show \eqref{eq:verif_form_neg} analogously by computing an upper bound on the network output.

\subsection{Bound Computation}
\label{sub:bound_comp}
In order to compute those bounds over the output of the network, verification algorithms require an oracle capable of concretizing linear functions over the input domain of the network, that is, optimisation of linear functions over the domain.
One easy way to provide this would be to decompose $\lzerodomain$ into an union of subdomains, where each subdomain would be defined by a possible sparsity pattern.
Each subdomain would then only be defined by independent inequalities, and the known concretization functions for interval bounds would be applicable. 
However, this would require a combinatorially large number of subdomains, with $\lzerodomain$ having to be split into $\binom{n}{l}$ subdomains. 

Instead, we derive in Algorithm~\ref{alg:lzero_conc} a concretization function for $\lzerodomain$ that is $\mathcal{O}(n)$:
\begin{equation}
\begin{aligned}
    \min \quad &\cb^T \xb \\
    \text{such that }& \| \xb \|_0 = l\\
    & x_i \geq 0 \implies \epsilon \leq x_i \leq 1
    \end{aligned}
\end{equation}

\begin{algorithm}
\caption{Concretization on $\lzerodomain$}
\label{alg:lzero_conc}
\begin{algorithmic}[1]

\Require $\cb \in \mathbb{R}^n$
\Ensure $\min \cb^T \xb$ such that $\xb \in \lzerodomain$

\State $\alphab \leftarrow \epsilon \cdot \cb^+ + \cb^-$ \\
where $c^+_i = \max\left(c_i, 0\right)$  and $c^-_i = \min\left(c_i, 0\right)$
\State $\texttt{coords} \leftarrow \texttt{top\_k}(-\alphab, l)$ \label{line:topk_step}

\State\Return $- \texttt{sum}(\texttt{coords})$
\end{algorithmic}
\end{algorithm}
The vector $\alphab$ represent, for each coordinate, the amount that it would contribute to the objective if it was included in the support. 
Step~\ref{line:topk_step} will pick the $l$ smallest contributions, in order to identify what the support should be. 
All the other coordinates which are not included in the support will have a value of zero, and are therefore not contributing anything to the objective function.

\subsection{Bound tightening}
\label{sub:tighten_bab}
This concretization function, combined with the backward linear relaxation based perturbation analysis algorithm~\citep{auto_lirpa}, allows us to compute bounds for all intermediate activations of the network, all the way up to the output of the network. 
However, there is no guarantee that the bounds computed this way will be sufficient to satisfy the requirements of specification~\eqref{eq:verif_props}. 
To obtain tighter bounds, we employ this bound computation as a sub-routine inside a branch-and-bound algorithm~\citep{bunel2020branch}.

\begin{figure}
    \centering
    \includesvg[width=\linewidth,inkscapelatex=false]{plots/ImpactOfPatterns.svg}
    \caption{
    Distribution of how far the bound is from being sufficient to prove Property~\eqref{eq:verif_form_pos} for $i=49$ on the $n=50,m=30$ network of Figure~\ref{fig:learned_50}. 
    Each dot correspond to a randomly chosen subdomain of $\left\{\xb \in \lzerodomain, x_{49} \geq \epsilon\right\}$. 
    The color indicates how many of the signal coordinates have been forced to be non-zero.
    \label{fig:impact_of_patterns}}
\end{figure}

For a given coordinate $i$, we sample 5000 subdomains of the specification~\eqref{eq:verif_form_pos} by imposing that a random selection of coordinates different of $i$ are either constrained to be greater than $\epsilon$ or constrained to be equal to $0$. 
The remaining coordinates are left constrained, beside the global requirements on the $\mathcal{L}_0$ norm.
For each subdomain, we count how many sparsity patterns are compatible with the sampled constraints, and we compute the lower bound produced by the bound propagation algorithm.
We observe in Figure~\ref{fig:impact_of_patterns} that the tightness of the bound is strongly correlated to the number of sparsity patterns.
As a result, we choose to use an input-branching strategy, rather than one branching on activations, as is most common in the adversarial robustness verification literature~\citep{wang2021beta}.
Input branching has the additional benefit of not requiring an inner-loop iterative algorithm to optimize the dual variables, which enables us to evaluate bounds on subdomains faster.

A simple branching strategy could divide the domain based on a coordinate $i$, imposing $x_i \geq \epsilon$ in one subdomain and $x_i = 0$ in the other.
However, this creates highly imbalanced branches, leading to inefficient verification. 
For instance, with $n=50$, and $\ell_0=5$, the original domain represents $\binom{50}{5} \approx 2.1\cdot10^9$.
The $x_i \geq \epsilon$ branch represents only $\binom{49}{4} \approx 2.1\cdot10^6$, while the $x_i = 0$ branch represents $\binom{49}{5} \approx 1.9 \cdot 10^9$ patterns. 
This imbalance significantly increases verification time.

We propose an alternative branching strategy more suited to the $\mathcal{L}_0$ bounded verification problems.
We identify subdomains of $\lzerodomain$ by an index $j \in [1, n]$ and a set of coordinates $\texttt{on}$.
\begin{equation*}
\begin{aligned}
        \xb \in \mathcal{D}(j, \texttt{on}, \lzerodomain) \iff & \xb \in \lzerodomain \\
        & \forall i \leq j,\ i \notin \texttt{on} \implies x_i = 0 \\
        & \ i \in \texttt{on} \implies x_i \geq \epsilon\\
    \end{aligned}
\end{equation*}
Given a subdomain $\mathcal{D}(j, \texttt{on})$, our branching function is\looseness=-1 
\begin{equation*}
    \texttt{split}\left(\mathcal{D}(j, \texttt{on})\right) = 
    \left\{\mathcal{D}(k, \texttt{on} \cup \{k\}) \ \forall k \in [j+1, n]\right\}
\end{equation*}
The index $j$ indicates that all coordinates up to the $j$-th one have been either included or excluded for the support.
The branching operation consists in choosing which coordinate is the next to be included in the support.
This branching heuristic does not make use of any information obtained during the bound computation, but has the advantage that each resulting subdomain is guaranteed to have at least one more active coordinate in the support determined.

During the branch and bound procedure, we may completely fix the sparsity pattern of a subdomain if $\texttt{length}(\texttt{on}) = l$ without being able to prove a sufficiently tight lower bound. 
A completely fixed sparsity pattern can be described using only interval constraints, and we can still continue to perform branching by picking the coordinate with the longest range, and splitting its domain in half, analogous to the longest edge heuristic of~\citet{bunel2018unified}.

We summarize the whole procedure in Algorithm~\ref{alg:lzero_bab}. 
The algorithm terminating means that for all the subdomains encountered, a lower bound of zero has been comptued, otherwise the subdomain would have been refined further (see line~\ref{line:loosebound}).
On the other hand, at each stage of the search procedure, we sample inputs from the subdomain (line ~\eqref{line:sample}), and evaluate our network on them.
If the output value is smaller than zero, that input is a counter-example to the property that we are trying to prove.

\begin{algorithm}
\caption{Branch-and-Bound Verification on $\lzerodomain$}
\label{alg:lzero_bab}
\begin{algorithmic}[1]

\Require Network parameters $\thetab$
\Require Coordinate $i$ for which we want to prove \eqref{eq:verif_form_pos}.
\Ensure Successful proof, or counterexample.

\State \texttt{domains} $\leftarrow \left[\mathcal{D}\left(0, \{i\}, \lzerodomain\right)\right]$

\While{\texttt{domains} is not empty}
  \State \texttt{d} $\leftarrow$ \texttt{domains.pop()}
  \State \texttt{lower} $\leftarrow$ \texttt{compute\_lower\_bound}($\thetab$, \texttt{d})
  \State \texttt{x\_test} $\leftarrow$ \texttt{sample}(\texttt{d})\label{line:sample}
  \State \texttt{out} $\leftarrow z_i(\texttt{x\_test}, \thetab)$ 
  \If{out $<$ 0}\Comment{Counterexample found.}\label{line:cexfound}
    \State \Return FALSE, \texttt{x\_test}
  \EndIf
  \If{\texttt{lower} $<$ 0}\Comment{Bound too loose.}\label{line:loosebound}
    \State \texttt{domains.extend(split(d))}
  \EndIf
\EndWhile
\State \Return True
\end{algorithmic}
\end{algorithm}

\subsection{Feature Comparison}
\label{feature:comp}
We present in Table~\ref{tab:feature_comp} a comparison of the benefits and inconvenient of our proposed method, as opposed to traditional approaches based on compressed sensing.

\begin{table*}
\centering
\begin{tabularx}{\linewidth}{p{.36\linewidth}>{\centering\arraybackslash}X>{\centering\arraybackslash}X}
\toprule
& \textbf{Traditional Solvers} & \textbf{Learned Decoders} \\
\midrule
\textbf{Inference Time} & \textcolor{orange}{Seconds} & \textcolor{green}{Milliseconds} \\
\midrule 
\textbf{Verification Time} & \textcolor{green}{0s (correct by construction)} & \textcolor{orange}{Seconds to hours} \\
\midrule
\textbf{New Sensing Matrix / Dimensions} & \textcolor{green}{Same algorithm apply.} & \textcolor{orange}{Retrain and verify each time.}\\ 
\midrule
\textbf{Novel Type of Measurements} & \textcolor{red}{New algorithm required.} & \textcolor{orange}{Simply change training data, adapt verification.} \\ \bottomrule          
\end{tabularx}
\caption{\label{tab:feature_comp}Feature comparison of Traditional Solvers versus Learned Decoders.}
\end{table*}

\begin{figure*}
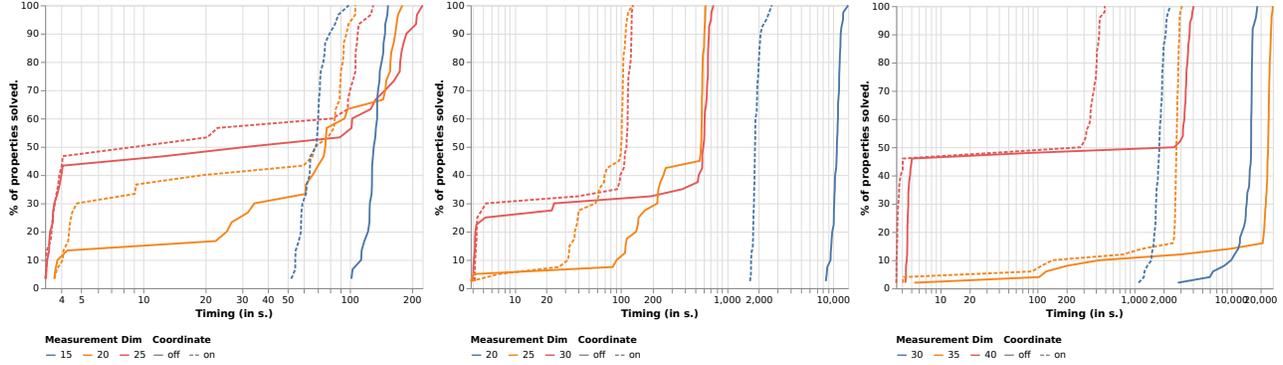

\centering
\begin{subfigure}{0.33\textwidth}
\centering
\includesvg[width=\textwidth,inkscapelatex=false]{plots/Learned_30.svg}
\caption{Recovering $n=30$ signals.}
\label{fig:learned_30}
\end{subfigure}%
\begin{subfigure}{0.33\textwidth}
\centering
\includesvg[width=\textwidth,inkscapelatex=false]{plots/Learned_40.svg}
\caption{Recovering $n=40$ signals.}
\label{fig:learned_40}
\end{subfigure}%
\begin{subfigure}{0.33\textwidth}
\centering
\includesvg[width=\textwidth,inkscapelatex=false]{plots/Learned_50.svg}
\caption{Recovering $n=50$ signals.}
\label{fig:learned_50}
\end{subfigure}
\caption{Cactus plots showing the impact of problem dimension on verification times. \textit{off} coordinates are much slower to verify than \textit{on} coordinates, and smaller number of measurements $m$ (blue or orange lines) are also associated with harder to verify properties. Increase in the signal size $n$ also influence the verification runtimes, which can be seen by comparing the scales of the time axis of the subfigures.}
\label{fig:comparison_problem_dimensions}
\end{figure*}

The main advantage of using a learned decoder lies in the inference time. 
Solving a problem amounts to executing a forward pass through the learned model, instead of applying an iterative algorithm.
As the network that we employ are small, evaluating them is very fast.
When operating on one signal at a time, for the blue network of Figure~\ref{fig:learned_50}, it took 0.27ms on GPU, or 2.1ms on CPU to identify the support of a signal from its measurement.
If we have multiple signals that we intend to recover at the same time, the learned decoder is also trivially amenable to batching.

Our proposed approach necessitates verifying the neural network correctness for a specific task.
This requirement is not present in classical algorithms, which are typically correct by design and provide assurances of converging to the correct solution. 
Although this verification step can be costly for learned models, it only needs to be performed once. 
For a given problem setting defined by the choice of sensing matrix and a learned model that claims to solve it, the verification procedure demonstrates that any possible measurements $\textbf{y}$ will be decoded correctly. 
Consequently, if numerous instances of the same problem need to be solved repeatedly, the potentially high cost of verification can be spread out over many samples.
This can be particularly advantageous when the latency until a problem solution is found is critical.
Section~\ref{sub:dimension_impact}, \ref{sub:sensing_mat} and \ref{sub:binary_constraint} dive deeper into the impact on verification time of different characteristics of the problem.

When the problem changes, whether due to a change in the $(n,m, \ell_0$) dimensions, or in the measurement process as defined by the sensing matrix, a new model must be trained and verified, which is not the case with classical algorithms.
On the other hand, if the measurement type changes, adapting the learned decoder only requires modifying the training pipeline and ensuring that bound propagation can still be performed.
This is easier than developing a new solving algorithm and proving its correctness.
We showcase this adaptability in Section~\ref{sub:binary_constraint} by demonstrating support for compressed sensing with binary measurement.

\section{Evaluation}
For each learned model, we verify the support identification properties~\eqref{eq:verif_props} of each of the dimension of the signal, using the method of Algorithm~\ref{alg:lzero_bab}, and measure the time required to prove correctness.
Unless otherwise specified, the verification is performed using branch-and-bound with the criterion described in Section~\ref{sub:tighten_bab}, and backward mode LiRPA bound propagation, using the concretization function described in Algorithm~\ref{alg:lzero_conc}.
All experiments are performed using the JaxVerify verification toolkit~\citep{jaxverify2020github} implemented in Jax~\citep{jax2018github}. \looseness=-1

\subsection{Impact of Dimension}
\label{sub:dimension_impact}
We vary both the dimension of the sparse signal $n$, the number of measurements $m$, for signals with a known pseudo-norm of the signal $\ell_0 = 5$.
For each configuration, we train a model capable of performing the recovery of the support, using the methodology described in Section~\ref{sub:at_for_pareps}. 
For each plot, all models have the same architecture.

All networks are completely verified, as shown by the fact that all curves of Figure~\ref{fig:comparison_problem_dimensions} reach the 100$\%$ mark.
In all settings, it is faster to verify properties of the form~\eqref{eq:verif_form_pos}, denoted \textit{on}, than~\eqref{eq:verif_form_neg}, denoted \textit{off}.
This can be explained by the influence of the number of the sparsity patterns covered by a domain on the tightness of the bound, as previously highlighted in Figure~\ref{fig:impact_of_patterns}.

We also observe large plateaus where to obtain an increase in the proportion of properties being verified, a large increase in verification runtime is required.
Some properties are verifiable with little to no branching, while others require a much finer decomposition, leading to the explosion in verification runtime.


Larger verification runtime appear to be caused by smaller number of measurements.
Our hypothesis is that with fewer measurements, the decoding function needs to be more complex and nonlinear.
In the limit of $m=n$, a linear decoding function would be sufficient.
LiRPA methods perform linear approximations of the network, so a less linear model leads to looser bounds and more branching steps for correctness verification. This is evident in Figure~\ref{fig:comparison_problem_dimensions} where red lines (fewer measurements) require more branching steps compared to blue/orange lines (more measurements).

\subsection{Impact of Learning the Sensing Matrix}
\label{sub:sensing_mat}
If we get to choose our sensing matrix, we can optimize it jointly with the decoder to make the problem easier to verify.
Otherwise, if it is pre-determined, for example because it is the result of a physical sensing process, we can keep it as a constant and optimize only the parameters $\Theta$ of the decoder.

To evaluate the ability of our training procedures to handle fixed sensing matrices, we train decoders for problems of same size, and compare the ease of performing verification in two settings: $\textit{Learned}$, similar to the configuration of the previous section, and $\textit{Fixed}$, with a randomly sampled matrix with gaussian iid entries.
This matrix is known at training time, but is kept fixed while we train the decoder.

\begin{figure}
    \centering
    \includesvg[width=\linewidth,inkscapelatex=false]{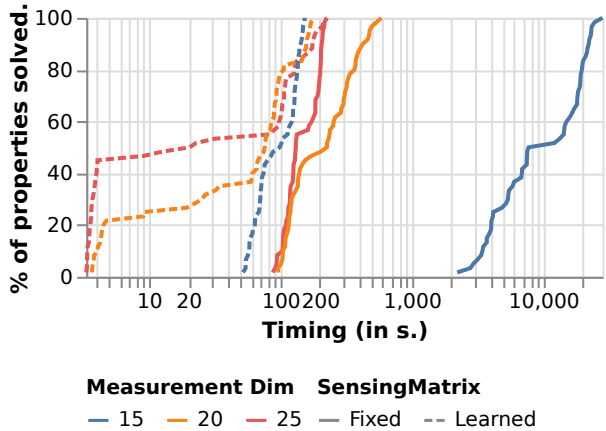}
    \caption{Impact of $\textit{Fixed}$ vs. $\textit{Learned}$ sensing matrix.
    Optimizing the sensing matrix leads to significantly faster verification, particularly for hard problems.}
    \label{fig:fixed_versus_learned}
\end{figure}

As can be seen in Figure~\ref{fig:fixed_versus_learned}, learning the sensing matrix can lead to significant reduction in the verification runtime.
Nevertheless, all models were able to be completely verified.
For the more difficult model (low number of measurements), the impact was more significant. \looseness=-1

\subsection{Incorporating additional constraints}
\label{sub:binary_constraint}
To demonstrate the transferability of our method to new settings, we train a network to solve a support identification problem for signals of size 15, using partially binarized measurements ($m_1 = 5$, $m_2=4$). 
The training procedure remains exactly the same, with only the computation of the measurements requiring changes.
To make verification feasible, we adapt the proof procedure to use the bound computation method of ~\citet{wang2021beta} and use branching over intermediate activations, such as the $\texttt{sign}$ operations binarizing the measurements. 

\begin{figure}
    \centering
    \includesvg[width=\linewidth,inkscapelatex=false]{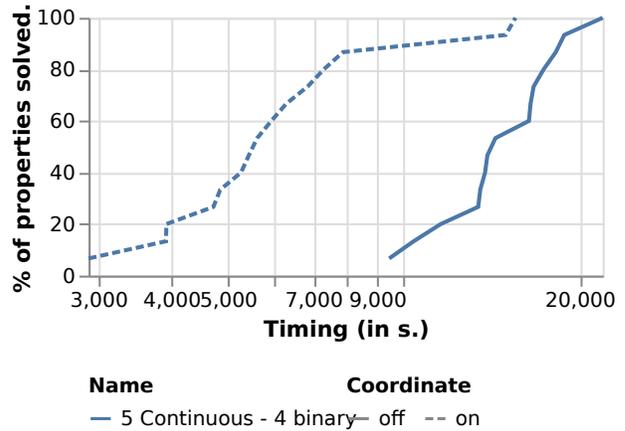}
    \caption{Runtime for correctness proofs of model handling binary measurements.}
    \label{fig:learned_binary}
\end{figure}

Despite the problem dimensions being quite small, Figure~\ref{fig:learned_binary} show relatively high verification runtimes.
This is the consequence of each bound computation being significantly more expensive, due to the necessity of optimizing dual variables.
Nevertheless, we manage to fully verify the learned model.
We also ensured that the problem was non-trivial and could not be solved by ignoring the binary measurements.

\subsection{Comparison with baseline}
We consider the problem of estimating signals in dimension $n=30$ given $m = 15$ measurements. Similar to the previous subsections, we set $\epsilon = 0.5$ and the $\ell_0$ norm $l = 5$. 
Our neural solver is provably correct in this setting even with a fixed sensing matrix.
Since the dimensionality of this setting is relatively small, we can solve a mixed integer program that attempts to find a counter-example ${\bf x}^*_b \in \lzerodomain$ where basis pursuit solution ${\bf x}_b$ is incorrect.
Indeed, the mixed integer program finds an example where $||A{\bf x}^*_b - A{\bf x}_b|| > 0.2$.
Note that basis pursuit does not incorporate the additional information that the minimum value of the non-zero elements of ${\bf x}$ is $\epsilon$ as this would result in non-convex constraints.
One could arguably design an appropriate convex relaxations for these constraints, thereby resulting in a better hand-designed baseline. 
However, this is precisely the advantage that we wish to highlight for our proposed approach. 
While traditional methods require onerous labour by experts in the field for every new setting of compressed sensing, our method simply requires a way to generate the relevant training data. \looseness=-1

\vspace{-6.0pt}
\section{Discussion}
\vspace{-6.0pt}
We show that the neural network verification can indeed verify provably correct neural decoders for compressed sensing problems, at least for moderate problem dimensions. We believe that we have just scratched the surface in terms of the potential for these approaches to develop verifiably correct novel algorithms for challenging mathematical and computational tasks. In particular, developing stronger verification algorithms tailored to mathematical problems and training algorithms that can regularize a network towards being easier to verify are two promising directions for future work.\looseness=-1

\bibliography{ref.bib}

\end{document}